\documentclass[twoside,11pt]{article}

\usepackage{blindtext}

\usepackage[abbrvbib,preprint]{jmlr2e}
\usepackage{bm,bbm,booktabs,pifont,graphicx,float,amsmath,dirtree,fontawesome}
\usepackage{xcolor}
\usepackage{subcaption}
\definecolor{markcolor}{HTML}{DDDDFF}

\newcommand{\folder}{\faFolder}

\usepackage{lastpage}
\jmlrheading{23}{2026}{1-\pageref{LastPage}}{1/21; Revised 5/22}{9/22}{21-0000}{Plasticine Team}

\ShortHeadings{Plasticine: Accelerating Research in Plasticity-Motivated Deep RL}{Yuan, Wang, Ma, Sun, Li, Jin, Wang, Yang, Zeng, Tao, and Chen}
\firstpageno{1}

\usepackage{graphicx}
\usepackage{wrapfig}
\usepackage{cite}
\usepackage{duckuments}
\usepackage{CJKutf8}
\usepackage{enumitem}
\usepackage{algorithm}  
\usepackage{algorithmicx}  
\usepackage{algpseudocode}  
\usepackage{overpic}
\usepackage{xcolor} 

\definecolor{MyDarkBlue}{rgb}{0,0.5,1}
\definecolor{MyDarkGreen}{rgb}{0.02,0.6,0.02}
\definecolor{MyDarkRed}{rgb}{0.8,0.02,0.02}
\definecolor{MyDarkOrange}{rgb}{0.40,0.2,0.02}
\definecolor{MyYellow}{rgb}{1,0.55,0}
\definecolor{MyPurple}{RGB}{111,0,255}
\definecolor{MyRed}{rgb}{1.0,0.0,0.0}
\definecolor{MyGold}{rgb}{0.75,0.6,0.12}
\definecolor{MyDarkgray}{rgb}{0.66, 0.66, 0.66}
\definecolor{default}{RGB}{0,0,0}

\newcommand\ti[1]{\textit{#1}}

\renewcommand{\eqref}[1]{Eq.~(\ref{#1})} 

\begin{document}

\title{Plasticine: Accelerating Research in Plasticity-Motivated Deep Reinforcement Learning}

\author{
\begin{center}
    Mingqi Yuan\textsuperscript{1}\thanks{These authors contributed equally.}, Qi Wang\textsuperscript{2}\footnotemark[1], Guozheng Ma\textsuperscript{4}\footnotemark[1], Caihao Sun\textsuperscript{5}\footnotemark[1], Bo Li\textsuperscript{1}, Xin Jin\textsuperscript{3},\\
    Yunbo Wang\textsuperscript{2}, Xiaokang Yang\textsuperscript{2}, Wenjun Zeng\textsuperscript{3}, Dacheng Tao\textsuperscript{4}, Jiayu Chen\textsuperscript{56}\\
    {\normalfont
    \textsuperscript{1}HK PolyU$\quad$\textsuperscript{2}SJTU$\quad$\textsuperscript{3}EIT, Ningbo$\quad$\textsuperscript{4}NTU$\quad$\textsuperscript{5}HKU$\quad$\textsuperscript{6}INFIFORCE
    }
\end{center}
}

\editor{My editor}

\maketitle

\begin{abstract} 
Developing lifelong learning agents is crucial for artificial general intelligence (AGI). However, deep reinforcement learning (RL) systems often suffer from plasticity loss, where neural networks gradually lose their ability to adapt during training. Despite its significance, this field lacks unified benchmarks and evaluation protocols. We introduce Plasticine, the first open-source framework for benchmarking plasticity optimization in deep RL. Plasticine provides single-file implementations of over 13 mitigation methods, 6 evaluation metrics, and learning scenarios with increasing non-stationarity levels from standard to continually varying environments. This framework enables researchers to systematically quantify plasticity loss, evaluate mitigation strategies, and analyze plasticity dynamics across different contexts. 
Our documentation, examples, and source code are available at \url{https://github.com/RLE-Foundation/Plasticine}.
\end{abstract}


\section{Introduction}

Developing agents that never stop learning and evolving in dynamic real-world environments represents a critical frontier for advancing artificial intelligence and a fundamental prerequisite for continuously improving systems~\citep{abel2023definition}.
Reinforcement learning (RL), with its inherent properties of learning from experience, continual online improvement, and iterative strategic exploration, emerges as the ideal framework for pursuing this objective~\citep{silver2025experience}.
Despite its promise, recent research has increasingly revealed a critical limitation: plasticity loss, a phenomenon in which neural networks in RL agents gradually lose their ability to adapt and incorporate new information as training progresses~\citep{dohare2024loss, klein2024plasticity}, thus significantly impeding the development of truly lifelong learning agents~\citep{lyle2024switching}.

This phenomenon emerges specifically when switching from classical tabular RL methods to deep neural architectures~\citep{klein2024plasticity}, which explains why it has only recently become a priority in the research community.
Additionally, while related to the stability-plasticity trade-off studied in continual (un)supervised learning scenarios, plasticity loss presents a more acute challenge for RL systems due to the pronounced non-stationarity in both data distributions and learning objectives, combined with the bootstrapping mechanisms inherent to RL algorithms~\citep{nikishin2022primacy}. 
Furthermore, compared to preventing catastrophic forgetting, maintaining plasticity is considerably more critical for developing truly lifelong learnable agents~\citep{abel2023definition}.

Despite the clear importance of mitigating plasticity loss in deep RL, this field lacks unified methodological frameworks and evaluation protocols.
This lack of established benchmarks hampers fair comparisons, methodological consistency, and precise problem formulation. Numerous fundamental questions remain inadequately addressed in current research: 
$\bullet$~How can we reliably measure and quantify plasticity loss through empirical means? 
$\bullet$~To what degree do existing mitigation approaches effectively resolve this issue? 
$\bullet$~In what ways does plasticity loss vary across different task domains and learning scenarios?
To address these gaps, we introduce \ti{Plasticine}, the first open-source benchmark for plasticity research in deep RL. 
The following section presents the framework architecture. 
Detailed method descriptions, experimental configurations, and results appear in the Appendices.

\section{Architecture of Plasticine}
\begin{figure*}[t]
    \centering
    \begin{minipage}{0.91\linewidth}
        \centering
        \subcaptionbox{Architecture of the Plasticine framework. \textit{Vis.} is short for Visual.
        \label{fig:architecture}
        }%
            {\includegraphics[width=\linewidth]{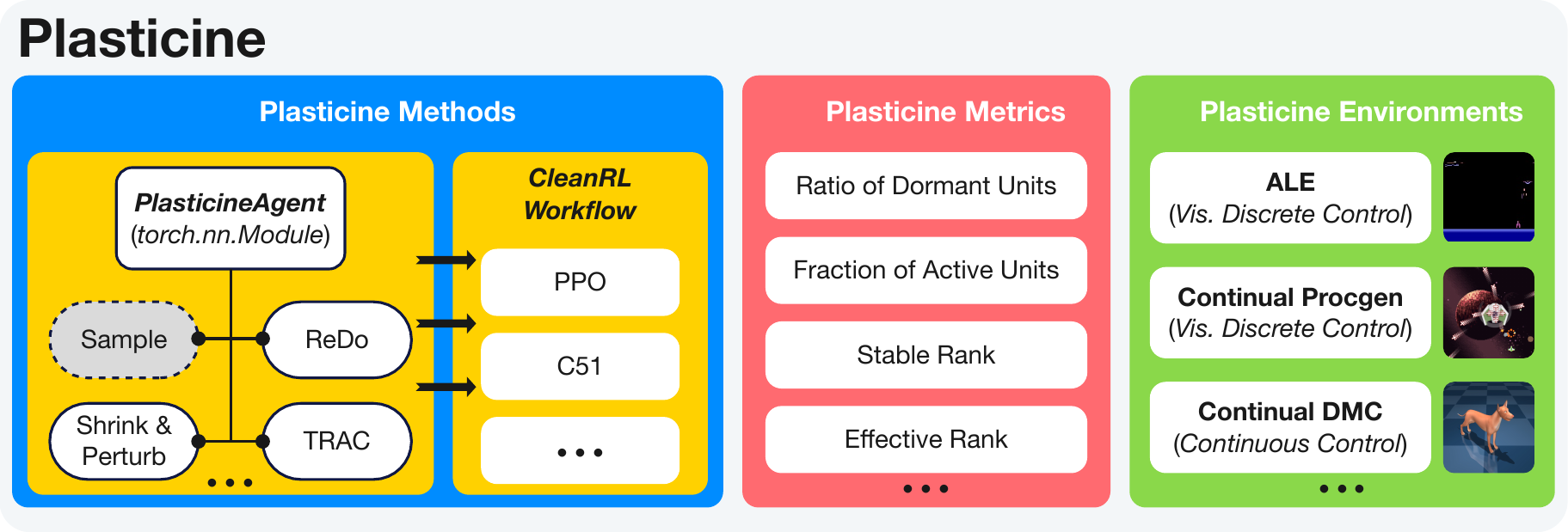}} %
        \begin{minipage}[b]{0.48\linewidth}
            \centering
            \subcaptionbox{Code implementation of the Plasticine-enhanced agent.
            \label{fig:code_implementation}
            }%
                {\includegraphics[width=\linewidth]{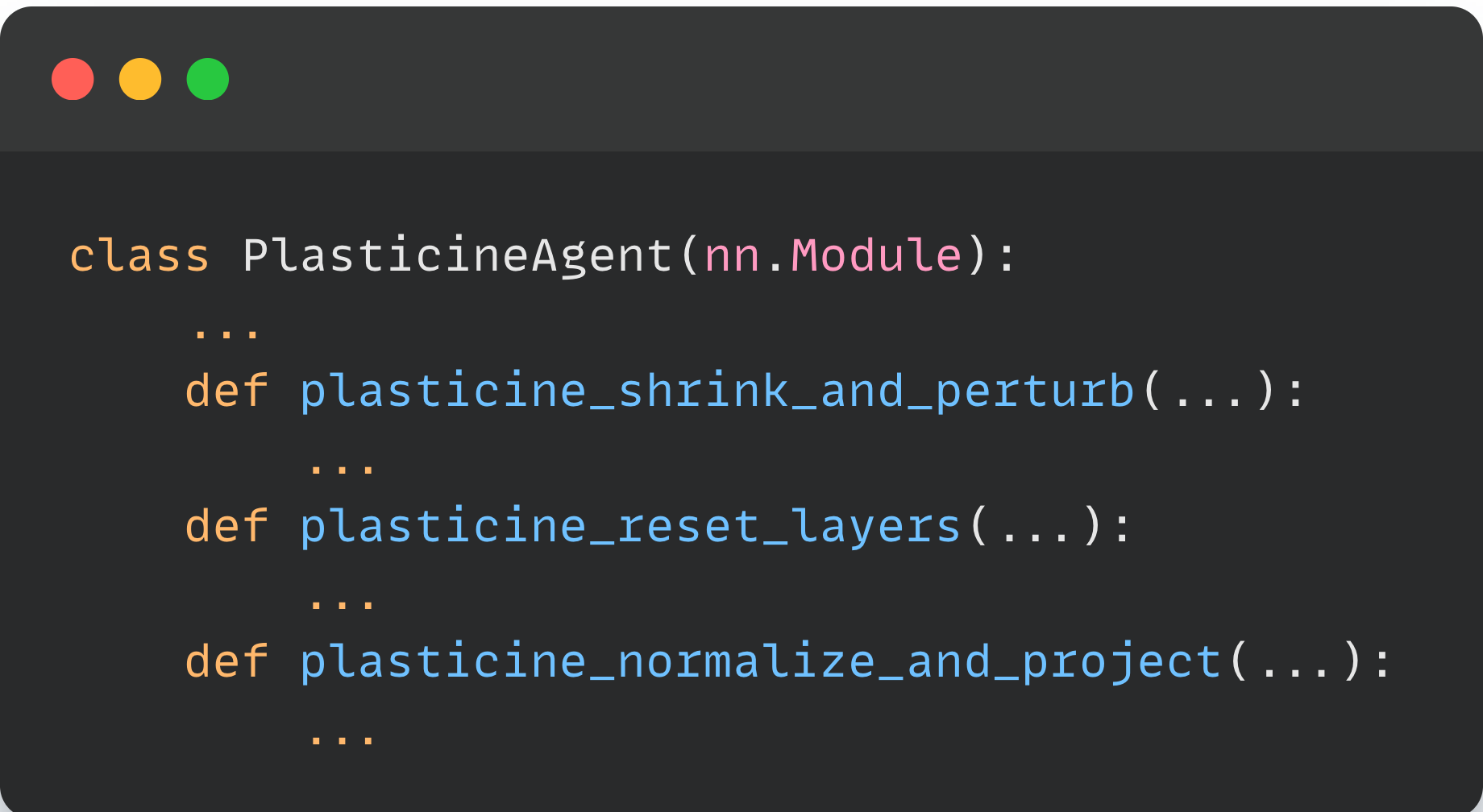}}
        \end{minipage}
        \hfill 
        \begin{minipage}[b]{0.48\linewidth}
            \centering
            \subcaptionbox{Invocation of the Plasticine methods in the \textit{CleanRL} workflow.}%
                {\includegraphics[width=\linewidth]{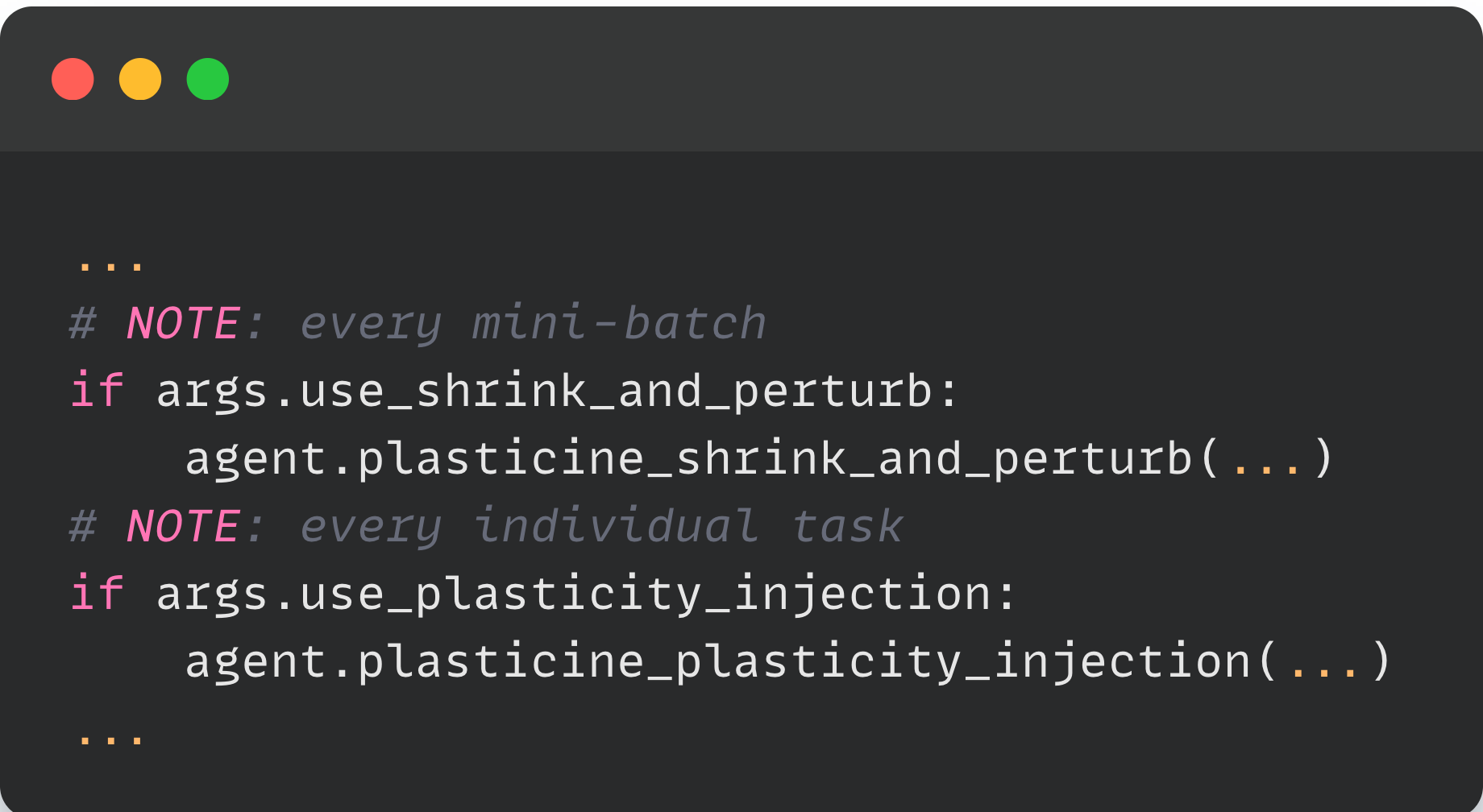}}
        \end{minipage}
    \end{minipage}
    \vspace{-7pt}
    \caption{Overview of the Plasticine framework. Plasticine strikes a balance between ``\textbf{single-file}" and ``\textbf{modularized}" implementations, facilitating related research and redevelopment while effectively controlling the project size.}
    \label{fig:overview}
    \vspace{-17pt}
\end{figure*}

Figure~\ref{fig:overview} illustrates the overall architecture of the \textit{Plasticine} framework, which contains three core parts: \textbf{Methods}, \textbf{Metrics}, and \textbf{Environments}. To maximize the convenience for researchers, we adopt the ``\textbf{single-file}'' implementation style that has been widely recognized by the RL community \citep{huang2022cleanrl,tarasov2023corl}. However, as most plasticity enhancement methods are relatively small techniques that intervene in the agent's network, adopting a purely single-file implementation style would lead to excessive project volume and management difficulties. Therefore, we strike a balance between single-file and modular design. As shown in Figure~\ref{fig:architecture} and \subref{fig:code_implementation}, all plasticity methods are integrated into the agent implementation, which is then combined with CleanRL's single-file RL algorithm \citep{huang2022cleanrl}. This approach allows researchers to focus on plasticity methods and facilitates redevelopment, while effectively controlling the project size. The following sections detail the design insights.

\subsection{Plasticity Loss Mitigation Methods}
Plasticity loss occurs when neural networks gradually develop optimization pathologies during non-stationary RL training processes, causing them to lose their inherent learnability~\citep{nikishin2024parameter}. Therefore, targeted improvements in various aspects of neural network design have the potential to mitigate plasticity loss. In this benchmark, we have categorized existing mitigation methods based on their primary focus and implemented representative methods for evaluation. The details of each method can be found in Appendix~\ref{appendix:po methods}.
\begin{enumerate}[itemsep=1pt, parsep=0.5pt, topsep=2pt]
\item \textbf{Reset-based Intervention}: Re-initializing portions of the network represents the most direct approach to regain plasticity~\citep{nikishin2022primacy}, whether through manually selecting specific layers or automatically identifying neurons experiencing severe performance plateaus~\citep{sokar2023dormant, galashov2024non}.
\item \textbf{Normalization}:
Normalization techniques, exemplified by Layer Norm, have been shown to be highly effective in mitigating plasticity loss~\citep{nauman2024bigger}. Building on this, more advanced normalization approaches have emerged targeting plastic training, including NaP~\citep{lyle2024normalization} and hyperspherical norm~\citep{lee2025hyperspherical}.
\item \textbf{Regularization}:
Constraining parameter norms or feature rank effectively prevents unbounded growth of weight/feature norms and avoids representation collapse, factors considered highly correlated with plasticity~\citep{kumar2023maintaining, lyle2022understanding}.
\item \textbf{Activation Function}:
Alternative activation functions beyond standard ReLU significantly impact neuron activity patterns, helping prevent dead neurons and maintain network capacity throughout non-stationary training~\citep{park2025activation}.
\item \textbf{Optimizer}:
Specialized optimization approaches like Fast TRAC~\citep{muppidi2024fast} and Kron~\citep{castanyer2025stable} help navigate complex loss landscapes during non-stationary training, thereby maintaining plasticity.
\end{enumerate}

\subsection{Comprehensive Evaluation Metrics}
Accurately quantifying neural network plasticity remains an open research question~\citep{lyle2024disentangling}. 
Current methodologies typically involve monitoring multiple plasticity-related indicators from various perspectives and analyzing them holistically~\citep{SimBa, ceron2023small}. 
Our benchmark incorporates key metrics directly into the training pipeline, enabling systematic comparison of plasticity dynamics across different approaches. 
These metrics include the ratio of dormant to active neurons, which indicates possible neuronal pathologies~\citep{sokar2023dormant, ma2024revisiting}; multiple feature rank measurements (such as Stable Rank and Effective Rank) that track representation capacity and potential collapse~\citep{lyle2022understanding}; and weight/gradient norm calculations that capture problems such as growth of unbounded parameters and vanishing gradients. 
This multidimensional evaluation framework helps researchers pinpoint which specific aspects of neural network behavior are most significantly influenced by different plasticity optimization methods.

\subsection{Progressive Evaluation Scenarios}

Plasticity loss represents a critical barrier to achieving truly lifelong RL applications, with its severity increasing proportionally to environmental non-stationarity. 
To comprehensively evaluate both the magnitude of plasticity loss and the effectiveness of mitigation methods, we provide a progressive framework of increasingly non-stationary task scenarios, enabling systematic analysis of how plasticity dynamics evolve across complexity thresholds and quantifying the preservation of learning capabilities as environmental demands intensify.

\subsubsection{Standard Online RL}
Standard online RL naturally exhibits non-stationarity as policies evolve and through bootstrapping processes. In these environments, agents must maintain plasticity throughout training to achieve optimal performance instead of being stuck at suboptimal levels. This setting offers a foundation for measuring plasticity during natural distribution shifts in policy optimization without introducing explicit task changes. Our baseline evaluation includes the Arcade Learning Environment (ALE;~\citealp{bellemare2013arcade}) as the benchmark, which has been increasingly employed to study plasticity loss in deep RL \citep{kumar2021implicit,kumar2022dr,sokar2023dormant,castanyer2025stable}.

\subsubsection{Continual RL}
Continual RL scenarios create more challenging plasticity demands by introducing explicit distribution shifts beyond those naturally occurring in standard RL. Following the practice in \citep{muppidi2024fast,tang2025mitigating}, we consider two scenarios: one in which visual features change within the same task, and another that requires adaptation across entirely different tasks. These environments test an agent's ability to maintain plasticity across both sudden task switches and gradual dynamics changes, providing a more challenging testbed than standard online RL. With clearly defined non-stationarity boundaries, we can directly observe how models behave before and after distribution shifts, gaining valuable insights into plasticity degradation patterns across different task distributions.

\section{Conclusion}
Maintaining plasticity is essential for building agents that learn continually throughout their lifetimes, a central goal as reinforcement learning enters the era of experience. We present Plasticine, an open-source framework designed to accelerate research in this direction. Plasticine offers three key advantages: (1) \textbf{ease of use} through single-file implementations built on familiar RL workflows; (2) \textbf{broad coverage} spanning on-policy and off-policy algorithms, state-based and vision-based domains, and evaluation scenarios from standard online RL to continual settings; and (3) \textbf{comprehensive analysis tools} for diagnosing plasticity dynamics across methods and environments. We will continue to expand Plasticine with new methods and environments, and hope it serves as a useful foundation for future research in this rapidly evolving field.

\clearpage\newpage

\appendix

\section{Supplementary Material}

We provide comprehensive supplementary materials to support the \textbf{Plasticine} framework:
\begin{itemize}
    \item Appendix~\ref{appendix:structure} lists the code file structure of the whole project;
    \item Appendix~\ref{appendix:po methods} provides the workflow and implementation details of all the implemented methods for plasticity loss mitigation;
    \item Appendix~\ref{appendix:metrics} provides formally mathematical definitions of the plasticity-related metrics;
    \item Appendix~\ref{appendix:bench} introduces the implementation details of the training environments for the two learning scenarios (standard and continual RL);
    \item Appendix~\ref{appendix:backbone} introduces the included backbone RL algorithms;
    \item Appendix~\ref{appendix:exps} provides experimental configurations and results.
\end{itemize}




\clearpage\newpage

\section{Project Structure}\label{appendix:structure}

\dirtree{%
.1 \folder $\!$ Plasticine/.
.2 \folder $\!$ plasticine/ \textcolor{magenta}{(Main algorithm implementation)}.
.3 ppo\_continual\_procgen\_base.py \textcolor{magenta}{(\textit{torch.nn.Module}-based agent integrated with diverse plasticity loss-mitigation techniques)}.
.3 ppo\_continual\_procgen\_plasticine.py \textcolor{magenta}{(Combination of the Plasticine agent and the CleanRL workflow)}.
.3 ....
.3 utils.py \textcolor{magenta}{(Auxiliary functions)}.
.3 ....
.2 \folder $\!$ plasticine\_envs/ \textcolor{magenta}{(Environments for algorithm evaluation)}.
.3 procgen\_wrappers.py \textcolor{magenta}{(Continual Procgen benchmark)}.
.3 dmc\_wrappers.py \textcolor{magenta}{(Continual DMC benchmark)}.
.3 ....
.2 \folder $\!$ plasticine\_metrics/ \textcolor{magenta}{(Plasticity-related evaluation metrics)}.
.3 metrics.py.
.3 ....
.2 \folder $\!$ requirements/ \textcolor{magenta}{(Dependencies for environments)}.
.3 requirements-ale.txt.
.3 requirements-prcogen.txt.
.3 ....
.2 \folder $\!$ scripts/ \textcolor{magenta}{(Training scripts)}.
.2 \folder $\!$ experimental/ \textcolor{magenta}{(Experimental code files)}.
.2 \folder $\!$ assets/ \textcolor{magenta}{(Miscellaneous files)}.
}

\clearpage\newpage

\section{Details of Implemented Plasticity Optimization Methods}\label{appendix:po methods}
\subsection{Reset-based Intervention}
\subsubsection{Shrink and Perturb}
Shrink and perturb (SnP) \citep{ash2020warm} periodically scales the magnitude of all weights in the network by a factor and then adds a small amount of noise. For a set of parameters $\Theta$, SnP adjusts them by 
\begin{equation}
    \Theta=\alpha\Theta_{\rm current}+\beta\Theta_{\rm init},
\end{equation} 
where $\alpha=1-\beta$ and $\Theta_{\rm init}$ is sampled from the initialization distribution of parameters.

We provide two options for the SnP operation: The user can choose to invoke it at specific intervals or after each gradient descent step (Soft SnP).

\subsubsection{Plasticity Injection}
Plasticity injection (PI) \citep{nikishin2023deep} replaces the final layer of a network with a new function that is a sum of the final layer’s output and the output of a newly initialized layer subtracted by itself. The gradient is then blocked in both the original layer and the subtracted new layer. In practice, the PI operation is only applied to the final layers of the network.

\subsubsection{ReDo}
This technique resets individual neurons within the network based on a dormancy criterion at fixed intervals \citep{sokar2023dormant}. At each intervention point, we check all the neurons in the network using a mini-batch of training data and reset the dormant neurons as defined in Eq.~(\ref{eq:rdu}).

\subsubsection{Resetting Layer}
This method involves periodically replacing the
weights of the layers in the network with newly initialized values \citep{nikishin2022primacy}. We also provide two options for this operation: At each intervention point, the user can choose to replace the weights of the final layers or all the layers in the network.

\subsection{Normlization}

\subsubsection{Layer Normalization}
A \colorbox{markcolor}{torch.nn.LayerNorm} function \citep{ba2016layer} is set before the activation at each layer. For convolutional layers, it is applied to the channel dimension.

\subsubsection{Normalize-and-Project}
The normalize-and-project (NaP) operation follows a two-step workflow: normalization and projection \citep{lyle2024normalization}. For normalization, layer normalization (or RMSNorm) is inserted before every nonlinearity in the network to stabilize pre-activation statistics and enable gradient mixing, which helps revive dormant ReLU units by correlating gradients across units. For projection, we periodically rescale the weights of each layer to their initial norms during training, decoupling parameter norm growth from the effective learning rate (ELR). This prevents ELR decay caused by unchecked norm growth while maintaining the benefits of normalization. Optionally, scale/offset parameters are either normalized jointly (for homogeneous activations) or regularized toward initial values to avoid drift. The process ensures stable training dynamics in nonstationary tasks without compromising performance on stationary benchmarks.

\subsection{Regularization}

\subsubsection{L2 Regularization}
This method \citep{lyle2023understanding} adds an additional loss term to the model update, which corresponds to the L2 norm of the weights of the network. This loss is scaled by a coefficient $\alpha$.

\subsubsection{Regenerative Regularization}
This method \citep{kumar2023maintaining} adds an additional loss term to the model update, which is the L2 norm of the difference between the current parameters of the network and the initial network parameters. This loss is scaled by a coefficient $\alpha$.

\subsubsection{Parseval Regularization}
This method \citep{chung2024parseval} adds an additional loss term to the model update. For each weight matrix $\mathbf{W}$, the Parseval regularization loss is defined as
\begin{equation}
    L_{\rm Parseval}(\mathbf{W})=\lambda\Vert\mathbf{W}\mathbf{W}^{\rm T}-s\mathbf{I}\Vert_{F},
\end{equation}
where $\lambda>0$ controls the regularization strength, $s>0$ is a scaling coefficient, $\mathbf{I}$ is the identity matrix of appropriate dimension and $\Vert\cdot\Vert_F$ denotes the Frobenius norm.

\subsection{Activation Function}

\subsubsection{CReLU Activation}
This method replaces ReLU activations with the CReLU activation function \citep{abbas2023loss}:
\begin{equation}
    f(x)=\text{Concatenate}(\text{ReLU}(x),\text{ReLU}(-x)),
\end{equation}
which ensures that the gradient is non-zero for all units in a given layer.

We concatenate the outputs of a linear layer by the feature dimension while concatenating the outputs of a convolution layer by the feature dimension.

\subsubsection{Deep Fourier Features}
This method \citep{lewandowski2025plastic} computes deep Fourier features. It replaces standard activation functions with concatenated sine and cosine activations 
\begin{equation}
    f(x)=\text{Concatenate}(\sin(x),\cos(x))
\end{equation}
in every layer to dynamically balance linearity and nonlinearity for sustained plasticity in continual learning.

Similarly, we concatenate the outputs of a linear layer by the feature dimension while concatenating the outputs of a convolution layer by the feature dimension.

\subsection{Optimizer}


%
\subsubsection{TRAC} 
TRAC \citep{muppidi2024fast} operates as a meta-algorithm layered on top of a base optimizer like gradient descent or Adam. At each time step, TRAC computes the policy gradient and uses the base optimizer to obtain a candidate update. Then, it applies a set of 1D tuners—each associated with a different discount factor—to evaluate how far to trust this candidate update versus staying near a reference point. These tuners use the Erfi potential function \citep{zhang2024discounted} to dynamically adjust a scaling factor based on past gradients, and their outputs are aggregated to determine the final weight update as a convex combination of the candidate and the reference point. This online and data-dependent mechanism ensures stability and adaptability without requiring manual tuning, enabling TRAC to generalize across varied tasks and distribution shifts while mitigating both mild and extreme forms of plasticity loss.

\subsubsection{KRON} 
 KRON \citep{castanyer2025stable} is a Kronecker-factored optimizer that addresses gradient pathologies in deep RL by providing curvature-aware updates. Unlike first-order optimizers like Adam that rely on diagonal preconditioning, Kron approximates the Fisher information matrix using Kronecker products, enabling more precise adaptation to the loss landscape's curvature. This is particularly crucial in non-stationary RL settings, where policy-dependent data distributions and bootstrapped targets lead to unstable optimization trajectories.

\clearpage\newpage

\clearpage\newpage
\section{Plasticity-Related Evaluation Metrics}\label{appendix:metrics}

\subsection{Ratio of Dormant Units}
Given an input distribution $\mathcal{D}$, let $h_i^{\ell}(x)$ denote the activation of neuron $i$ in layer $\ell$ under input $x \in \mathcal{D}$ and $H^{\ell}$ be the number of neurons in layer $\ell$. We define the score of a neuron $i$ (in layer $\ell$) via the normalized average of its activation as follows \citep{sokar2023dormant}:
\begin{equation}\label{eq:rdu}
    s_i^{\ell} = \frac{\mathbb{E}_{x \in \mathcal{D}}|h_i^{\ell}(x)|}{\frac{1}{H^{\ell}}\sum_{k=1}^{H^{\ell}}\mathbb{E}_{x \in \mathcal{D}}|h_k^{\ell}(x)|}
\end{equation}
We say neuron $i$ in layer $\ell$ is $\tau$-\textbf{dormant} if $s_i^{\ell} \leq \tau$. Then, the ratio of dormant units (RDU) is defined as the percentage of dormant neurons among all the neurons.

\subsection{Fraction of Active Units}
Although the complete mechanisms underlying plasticity loss remain unclear, a reduction in the number of active units within the network has been identified as a principal factor contributing to this deterioration. Hence, the fraction of active units (FAU) is widely used as a metric for measuring plasticity. Specifically, the FAU for neurons located in module $\mathcal{M}$, is formally defined as \citep{ma2024revisiting}:
\begin{equation}
    \text{FAU}(\mathcal{M})=\frac{\sum_{n\in\mathcal{M}}\mathbbm{1}(a_{n}(x)>0)}{N},
\end{equation}
where $a_{n}(x)$ represent the activation of neuron $n$ given the input $x$, and $N$ is the total number of neurons within module $\mathcal{M}$.

\subsection{Stable Rank}
The stable rank (SR) assesses the rank of the feature matrix, reflecting the diversity and richness of the representations learned by the network \citep{kumar2022dr}. Denote by $\mathbf{F}\in\mathbb{R}^{n\times m}$ the feature matrix with $n$ samples and $m$ features, $\sigma_{k}$ are the singular values sorted in descending order for $k=1,\dots,q$ and $q=\min(n,m)$, the SR is defined as 
\begin{equation}
    \text{SR}(\mathbf{F})=\min\{k:\frac{\sum_i^k\sigma_i}{\sum_j^q\sigma_j}>0.99\},
\end{equation}

\subsection{Effective Rank}
In the case of neural networks, the effective rank of a hidden layer measures the number of units that can produce the output of the layer \citep{roy2007effective}.
If a hidden layer has a low effective rank, then a small number of units
can produce the output of the layer, meaning that many of the units in
the hidden layer are not providing any useful information. Similar to SR, the effective rank (ER) is defined as
\begin{equation}
    \text{ER}(\mathbf{F})=\exp(H(p_1,\dots,p_q)),H(p_1,\dots,p_q)=-\sum_{k=1}^{q}p_{k}\log p_{k},
\end{equation}
where $q=\min(n,m)$ and $p_k=\frac{\sigma_k}{\Vert\bm{\sigma}\Vert_1}$.  



\subsection{Weight Difference}
To quantify how much a model changes during training or across different tasks in continual reinforcement learning, we compute the weight difference between two model states. This metric is useful for evaluating plasticity and adaptation, as well as detecting abrupt or minimal updates in parameter space.

\subsection{Gradient Norm}

The gradient norm (GN) quantifies the magnitude of gradients during optimization and reflects the strength of the learning signal \citep{abbas2023loss}. In continual reinforcement learning, tracking the gradient norm is essential for diagnosing issues such as vanishing gradients, overfitting, or under-training. Moreover, it provides insights into how much the model updates in response to new data or tasks.

Given a model parameter set $\boldsymbol{\theta} = \{\theta_1, \dots, \theta_P\}$ with corresponding gradients $\nabla \mathcal{L}(\theta_p)$, the L2 norm of the gradients is defined as:
\begin{equation}
    \text{GN} = \left( \sum_{p=1}^{P} \left\| \nabla \mathcal{L}(\theta_p) \right\|_2^2 \right)^{1/2}.
\end{equation}




\clearpage\newpage

\section{Benchmark Selection}\label{appendix:bench}
\subsection{ALE}
\begin{figure*}[h!]
    \centering
    \includegraphics[width=\linewidth]{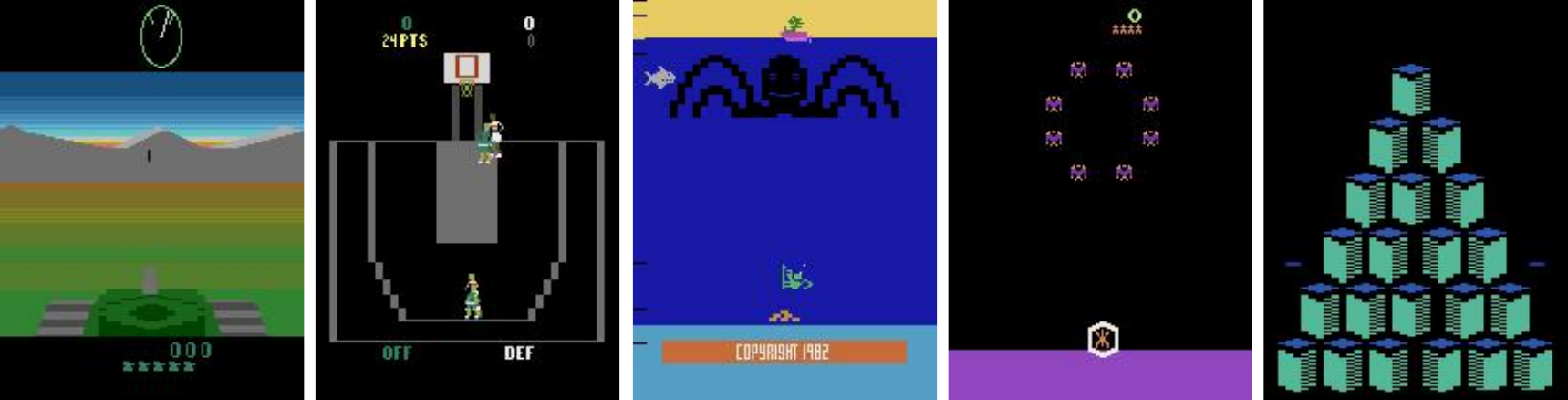}
    \caption{The ALE benchmark with visual observations and discrete action space.}
    \label{fig:ale_screenshots}
\end{figure*}

We first introduce the Arcade Learning Environment (ALE) as the benchmark, which is a collection of arcade game environments that require the agent to learn motor control directly from image-based observations. This benchmark has been used to examine the loss of plasticity in a series of continual RL studies \citep{kumar2021implicit,kumar2022dr,sokar2023dormant,castanyer2025stable}. As shown in Figure~\ref{fig:ale_screenshots}, the ALE environments often integrate distinct levels or stages. The transitions between these levels introduce significant changes in visual appearance, dynamics, or task objectives. Such level-wise shifts further exacerbate the non-stationarity of the learning problem, thereby leading to a significant challenge in terms of plasticity loss. In our experiments, we stack four consecutive frames to form an input state with the data shape of $(84, 84, 4)$ for all environments. 

\subsection{Continual Procgen}

\begin{figure*}[h!]
    \centering
    \includegraphics[width=\linewidth]{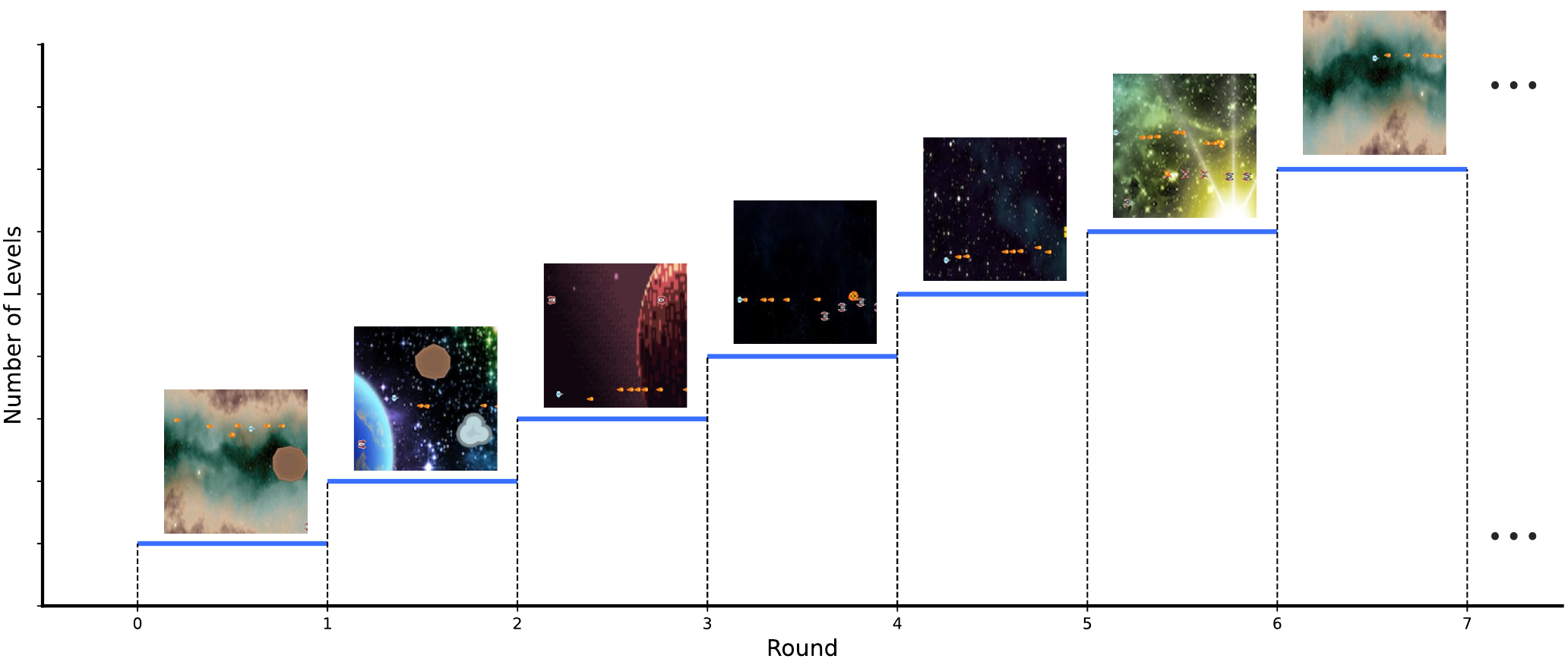}
    \caption{Continual Procgen benchmark with visual observations and discrete action space.}
    \label{fig:cont_pg_ls}
\end{figure*}

Furthermore, we introduce the Procgen benchmark, which comprises 16 procedurally generated environments. Procgen is developed similarly to the ALE benchmark, in which agents must learn motor control directly from image-based observations. However, Procgen involves much higher dynamics that require the agent to continually learn and adapt to the environment. Following the practice in \citep{muppidi2024fast} and \citep{tang2025mitigating}, we implement a variant of Procgen to support the comprehensive evaluation of the implemented methods. As shown in Figure~\ref{fig:cont_pg_ls}, distribution shifts are introduced by sampling a new procedurally generated level of the current game every 2 million time steps, and each level is considered a distinct task. In our experiments, all environments use a discrete fifteen-dimensional action space and generate $(64, 64, 3)$ RGB observations, with the \textit{easy} mode employed.

\subsection{Continual DMC}

\begin{figure*}[h!]
    \centering
    \includegraphics[width=\linewidth]{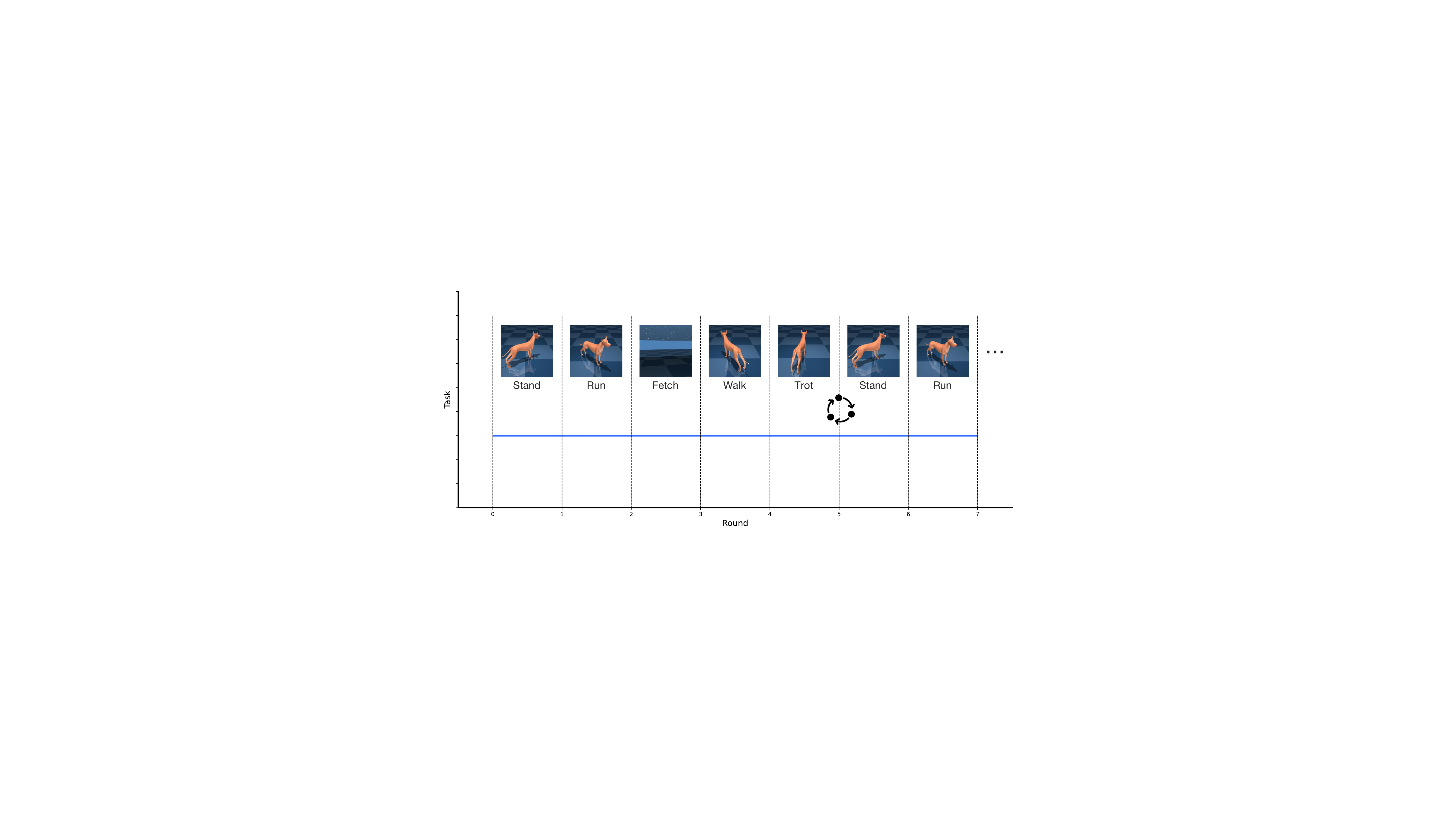}
    \caption{Continual DMC benchmark with proprioceptive observations and continuous action space.}
    \label{fig:cont_dmc_ts}
\end{figure*}

In addition to the discrete-action settings above, we further implement a variant of the DeepMind Control Suite (DMC) benchmark \citep{tassa2018deepmind} following \citep{tang2025mitigating}, which contains a series of continuous control tasks. Notably, each environment is a continual RL scenario that chains the
corresponding individual tasks in a sequence. As illustrated in Figure~\ref{fig:cont_dmc_ts}, the \textit{Continual Dog} environment chains four individual tasks: \textit{Dog Stand}, \textit{Dog Walk}, \textit{Dog Run}, and \textit{Dog Trot}. For each individual task, we train it for 1M environment steps.

\clearpage\newpage

\section{Backbone RL Algorithms}\label{appendix:backbone}
In this paper, we select two representative RL baselines as the backbone algorithms, which span policy-based, value-based, on-policy, and off-policy approaches.

\subsection{Proximal Policy Optimization}
Proximal policy optimization (PPO) \citep{schulman2017proximal} is a popular on-policy algorithm for learning a continuous or discrete control policy $\pi_{\bm{\theta}}(\bm{a}|\bm{s})$. 
PPO employs policy gradients using action-advantages: $A_t = A^\pi(\bm{a}_t, \bm{s}_t) = Q^\pi(\bm{a}_t, \bm{s}_t) - V^\pi(\bm{s}_t)$, and minimizes a clipped-ratio loss over mini-batches of recent experience (collected under $\pi_{\bm{\theta}_{\text{old}}}$):

\begin{equation}
L_\pi(\bm{\theta}) = -\mathbb{E}_{\tau \sim \pi} \left[ \min\left( \rho_t(\bm{\theta})A_t, \text{clip}(\rho_t(\bm{\theta}), 1-\epsilon, 1+\epsilon)A_t \right) \right],
\rho_t(\bm{\theta}) = \frac{\pi_{\bm{\theta}}(\bm{a}_t|\bm{s}_t)}{\pi_{\bm{\theta}_{\text{old}}}(\bm{a}_t|\bm{s}_t)}.
\end{equation}
Our PPO agents learn a state-value estimator---$V_{\bm{\phi}}(\bm{s})$, which is regressed against a target of discounted returns and used with generalized advantage estimation~\citep{schulman2015high}:

\begin{equation}
L_V(\bm{\phi}) = \mathbb{E}_{\tau \sim \pi} \left[ \left( V_{\bm{\phi}}(\bm{s}_t) - V_t^{\text{target}} \right)^2 \right].
\end{equation}

\subsection{Categorical DQN}

Categorical DQN (C51) \citep{bellemare2017distributional} is a distributional RL algorithm that models the return distribution rather than its expectation.
Instead of directly learning the action-value function $Q^{\pi}_{r}(\bm{s},\bm{a})$, C51 learns a categorical distribution of the random return $Z(\bm{s},\bm{a})$ on a fixed support $\{z_i\}_{i=1}^N$, where 


\begin{equation}
z_i = V_{\min} + (i-1)\Delta z,\Delta z = \frac{V_{\max}-V_{\min}}{N-1},\ \ i=1,\dots,N.
\end{equation}



The probability of atom $z_i$ is
$p_{\bm{\theta}}(z_i\mid \bm{s},\bm{a})=
\frac{\exp(f_i(\bm{s},\bm{a};\bm{\theta}))}{\sum_j \exp(f_j(\bm{s},\bm{a};\bm{\theta}))}$.
Here, $f_i(\bm{s},\bm{a};\bm{\theta})$ denotes the logit for atom $z_i$, $\bm{\theta}$ is a parametric model.
Given a sampled transition $(\bm{s},\bm{a},r,\bm{s'})$, C51 forms the distributional Bellman target using the target network $\bm{\theta}^-$,
then projects it back onto the fixed support via $\Phi$.
We minimize the Kullback–Leibler~(KL) divergence between the projected target and the predicted distribution:
\begin{equation}
L(\bm{\theta})=D_{\mathrm{KL}}\!\left(\Phi \hat{\mathcal{T}} Z_{\bm{\theta}^-}(\bm{s},\bm{a})
\ \|\ Z_{\bm{\theta}}(\bm{s},\bm{a})\right)= -\sum_{i=1}^N m_i \log p_{\bm{\theta}}(z_i\mid \bm{s},\bm{a}),
\end{equation}
where $m=\Phi \hat{\mathcal{T}} Z_{\bm{\theta}^-}(\bm{s},\bm{a})$, and $\hat{\mathcal{T}}$ denotes the distributional Bellman optimality operator.

\clearpage\newpage

\section{Experiments}\label{appendix:exps}
In this section, we conduct benchmark experiments for all three learning scenarios and compare the performance of the implemented approaches. The details of these experimental environments can be found in Appendix~\ref{appendix:bench}. The following figures only show a part of the experiment results, and more detailed benchmark experiment results can be found in Plasticine's Wandb space\footnote{https://wandb.ai/yuanmingqi/Plasticine/reportlist}.

\subsection{Plasticine+C51+ALE}

\begin{table}[htbp]
\centering
\begin{tabular}{llll}
\toprule
\textbf{Hyperparameter}       & \textbf{Value} & \textbf{Hyperparameter}   & \textbf{Value} \\ \midrule
Number of workers        & 1              & Number of atoms ($N$)           & 51             \\
Environments per worker  & 1              & Support lower bound ($V_{min}$) & -10            \\
Total timesteps          & 1e7            & Support upper bound ($V_{max}$) & 10             \\
Learning rate            & 2.5e-4         & Replay buffer size              & 1e6            \\
Discount factor          & 0.99           & Batch size                      & 32             \\
Target network frequency & 10000          & Start epsilon ($\epsilon$)      & 1              \\
Learning starts          & 80000          & End epsilon ($\epsilon$)        & 0.01           \\
Train frequency          & 4              & Exploration fraction            & 0.10           \\ \bottomrule
\end{tabular}
\caption{Hyperparameters used for the C51 agent in the standard ALE experiments.}
\end{table}

\begin{figure*}[h!]
    \centering
    \includegraphics[width=0.9\linewidth]{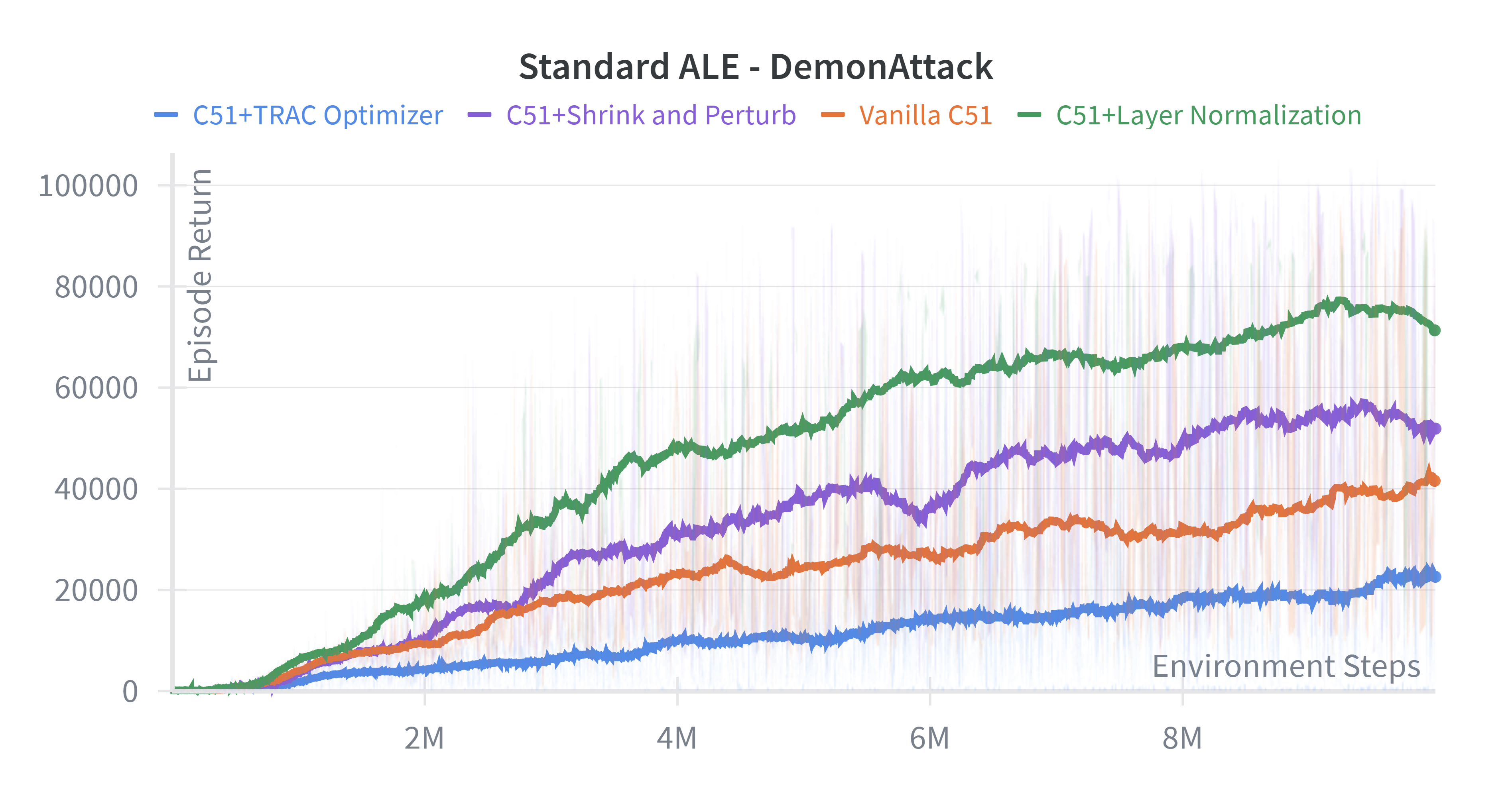}
    \caption{Performance comparison of the vanilla C51 agent and its combinations with several Plasticine methods on the \textit{ALE - DemonAttack} environment. The solid line and shaded region represent the mean and standard deviation across multiple runs, respectively.}
    \label{fig:c51_ale_demonattack}
\end{figure*}

\clearpage\newpage

\subsection{Plasticine+PPO+Continual Procgen}

\begin{table}[h!]
\centering
\begin{tabular}{llll}
\toprule
\textbf{Hyperparameter}       & \textbf{Value} & \textbf{Hyperparameter} & \textbf{Value} \\ \midrule
Number of tasks               & 10             & Level offset            & 20             \\
Observation downsampling      & (84, 84)       & Optimizer               & Adam (default)          \\
Observation normalization     & / 255.         & Learning rate           & 1e-3           \\
Reward normalization          & Yes            & GAE coefficient         & 0.95           \\
LSTM                          & No             & Entropy coefficient     & 0.01           \\
Stacked frames                & 4              & Value loss coefficient  & 0.5            \\
Environment steps (per round) & 2e6            & Value clip range        & 0.2            \\
Episode steps                 & 1000           & Max gradient norm       & 0.5            \\
Number of workers             & 1              & Number of mini-batches  & 8              \\
Environments per worker       & 1              & Discount factor         & 0.99           \\ \bottomrule
\end{tabular}
\caption{Hyperparameters used for the PPO agent in the continual procgen experiments.}
\end{table}

\begin{figure*}[h!]
    \centering
    \includegraphics[width=\linewidth]{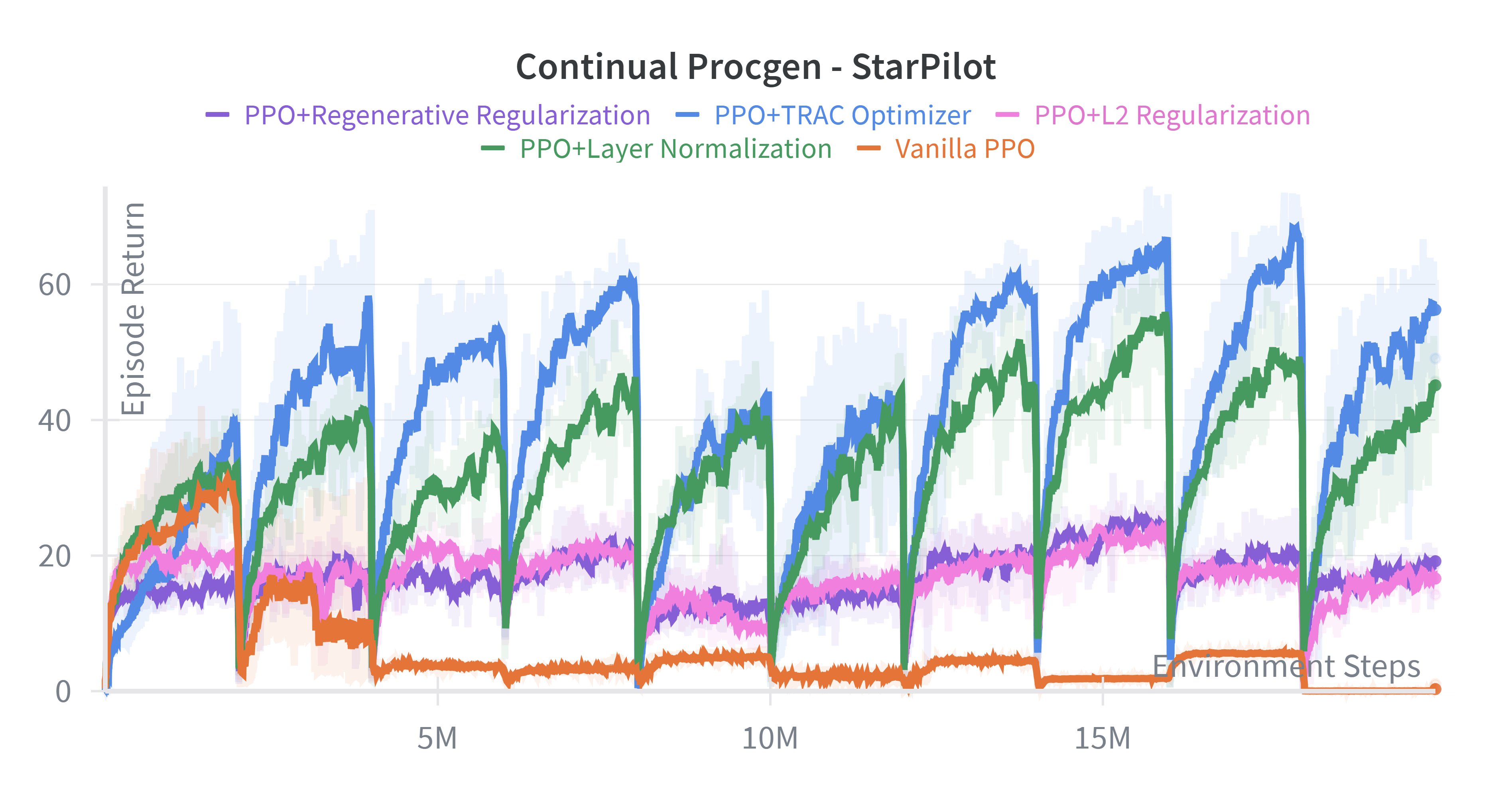}
    \caption{Performance comparison of the vanilla PPO agent and its combinations with several Plasticine methods on the \textit{Continual Procgen - StarPilot} environment. The solid line and shaded region represent the mean and standard deviation across multiple runs, respectively.}
    \label{fig:ppo_cont_procgen_starpilot}
\end{figure*}

\clearpage\newpage

\subsection{Plasticine+PPO+Continual DMC}

\begin{table}[h!]
\centering
\begin{tabular}{llll}
\toprule
\textbf{Hyperparameter}       & \textbf{Value} & \textbf{Hyperparameter} & \textbf{Value} \\ \midrule
Learning rate                 & 3e-4           & Optimizer               & Adam (default) \\
Reward normalization          & Yes            & GAE coefficient         & 0.95           \\
LSTM                          & No             & Entropy coefficient     & 0.0            \\
Initial std.                  & 1.0            & Value loss coefficient  & 0.5            \\
Environment steps (per round) & 1e6            & Value clip range        & 0.2            \\
Episode steps                 & 2048           & Max gradient norm       & 0.5            \\
Number of workers             & 1              & Number of mini-batches  & 32             \\
Environments per worker       & 1              & Discount factor         & 0.99           \\ \bottomrule
\end{tabular}
\caption{Hyperparameters used for the PPO agent in the continual DMC experiments.}
\end{table}

\begin{figure*}[h!]
    \centering
    \includegraphics[width=\linewidth]{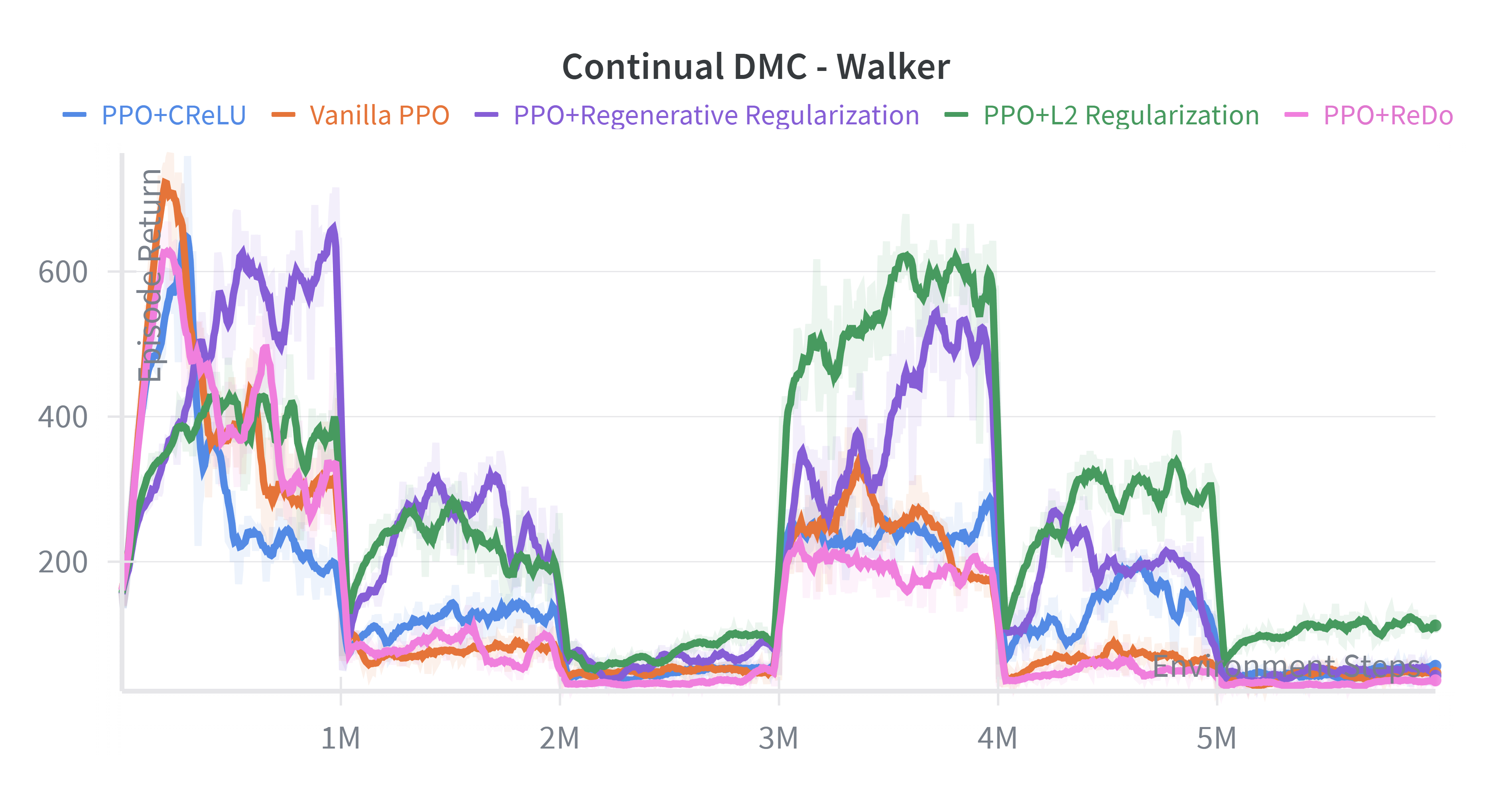}
    \caption{Performance comparison of the vanilla PPO agent and its combinations with several Plasticine methods on the \textit{Continual DMC - Walker} environment. The solid line and shaded region represent the mean and standard deviation across multiple runs, respectively.}
    \label{fig:ppo_cont_dmc_walker}
\end{figure*}

\clearpage\newpage

\section*{Acknowledgements}\label{appendix:ack}
This work is supported, in part, by the Hong Kong SAR Research Grants Council under Grant No. PolyU 15224823, the Guangdong Basic and Applied Basic Research Foundation under Grant No. 2024A1515011524, the NSFC under Grant No. 62302246, the ZJNSFC under Grant No. LQ23F010008, the Ningbo under Grants No. 2023Z237 \& 2023CX050011 \& 2024Z284 \& 2024Z289 \& 2025Z038 \& 2025Z059, and the Ningbo Institute of Digital Twin (IDT) under Grant No. S203.2.01.32.002. We thank the high-performance computing center at INFIFORCE Intelligent Technology Co., Ltd., Eastern Institute of Technology, and IDT for providing the computing resources.

\bibliography{sample}

\end{document}